\documentclass[runningheads]{llncs}
\usepackage{graphicx}
\usepackage{comment}
\usepackage{amsmath,amssymb} % define this before the line numbering.
\usepackage{color}

\usepackage{paralist}
\usepackage{booktabs}
\usepackage{xspace} % for \eg, \ie, etc.
\usepackage{stmaryrd}
\usepackage[normalem]{ulem}
\usepackage[tableposition=top]{caption}
\usepackage{appendix}

\usepackage{color}
\definecolor{crimson}{rgb}{0.86, 0.08, 0.24}
\definecolor{gray}{rgb}{0.5,0.5,0.5}
\definecolor{green}{rgb}{0, 0.4, 0}
\definecolor{orange}{rgb}{1, 0.5, 0}
\definecolor{mahogany}{rgb}{0.75, 0.25, 0.0}
\definecolor{purple}{rgb}{0.6, 0, 0.6}
\definecolor{darkgreen}{rgb}{0, 0.4, 0}
\definecolor{frenchblue}{rgb}{0.0, 0.45, 0.73}
\definecolor{red}{rgb}{1,0,0}
\definecolor{yellow}{rgb}{1,1,0}
\definecolor{magenta}{rgb}{1,0,1}
\definecolor{pink}{rgb}{1,0.412,0.706}

\newcommand{\mycomment}[1]{}

\newcommand{\rpm}{\raisebox{.2ex}{$\scriptstyle\pm$}}

\makeatletter
\DeclareRobustCommand\onedot{\futurelet\@let@token\@onedot}
\def\@onedot{\ifx\@let@token.\else.\null\fi\xspace}

\def\eg{\emph{e.g}\onedot}

\makeatother

\def\eg{e.g.,~}               % for example
\newlength\paramargin
\newlength\figmargin
\newlength\subfigmargin
\newlength\secmargin
\newlength\subsecmargin
\newlength\tabmargin
\newlength\eqmargin

\newcommand{\secref}[1]{Section~\ref{sec:#1}}
\newcommand{\subsecref}[1]{Section~\ref{subsec:#1}}
\newcommand{\figref}[1]{Figure~\ref{fig:#1}} 
\newcommand{\tabref}[1]{Table~\ref{tab:#1}}

\long\def\ignorethis#1{}

\newcommand{\tb}[1]{\textbf{#1}}

\usepackage[pagebackref=true,breaklinks=true,letterpaper=true,colorlinks,bookmarks=false,citecolor=blue,linkcolor=blue,urlcolor=blue,]{hyperref}

\usepackage[width=122mm,left=12mm,paperwidth=146mm,height=193mm,top=12mm,paperheight=217mm]{geometry}

\def \submission {}
\ifx \submission \undefined
	\newcommand{\charles}[1]{{\color{frenchblue}{#1}}}

\else
	\newcommand{\charles}[1]{{#1}}

\fi

\begin{document}
% \linenumbers
\pagestyle{headings}
\mainmatter
\def\ECCVSubNumber{169}  % Insert your submission number here

\title{Controllable Image Synthesis via SegVAE} % Replace with your title

\begin{comment}
\titlerunning{ECCV-20 submission ID \ECCVSubNumber} 
\authorrunning{ECCV-20 submission ID \ECCVSubNumber} 
\author{Anonymous ECCV submission}
\institute{Paper ID \ECCVSubNumber}
\end{comment}
%******************

% \begin{comment}
\titlerunning{Controllable Image Synthesis via SegVAE}
\author{Yen-Chi Cheng$^{1,2}$, Hsin-Ying Lee$^1$, Min Sun$^2$, Ming-Hsuan Yang$^{1,3}$}
\institute{$^1$University of California, Merced\hspace{5pt}$^2$National Tsing Hua University\hspace{5pt}$^3$Google Research}

\authorrunning{Y.-C. Cheng, H.-Y. Lee, M. Sun, M.-H. Yang}

% \end{comment}
%******************
\maketitle

% teaser
\begin{figure}[th]
\vspace{-5mm} 
\centering
\includegraphics[width=\linewidth]{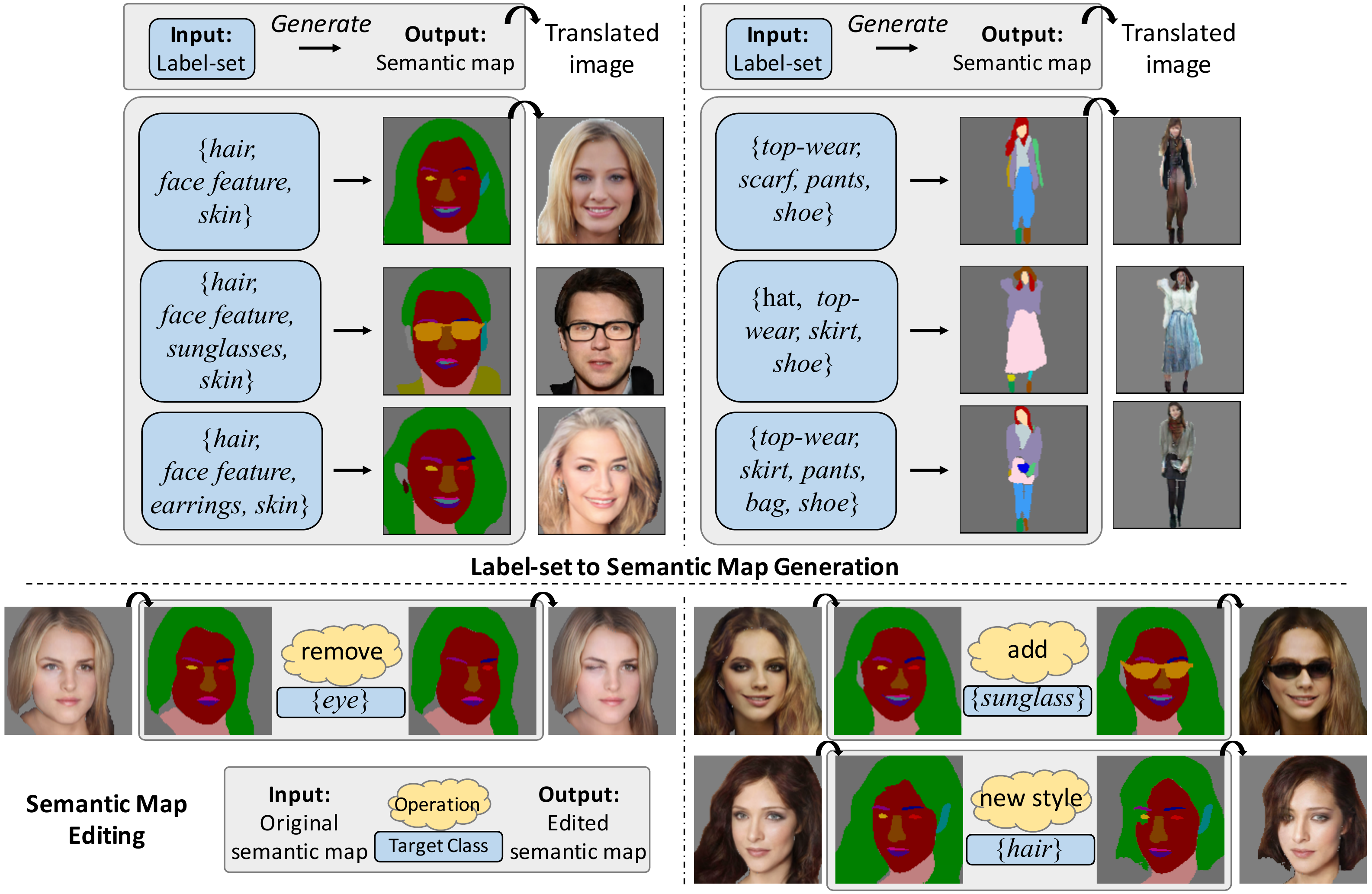}
\caption{
    \textbf{Label-set to Semantic map generation.}
    (\textit{Top}) Given a label-set, our model can generate diverse and realistic semantic maps. Translated RGB images are shown to better visualize the quality of the generated semantic maps.
    (\textit{Bottom}) The proposed model enables several real-world flexible image editing.
    }
\vspace{-3mm}
\end{figure}

\begin{abstract}
Flexible user controls are desirable for content creation and image editing.
A semantic map is commonly used intermediate representation for conditional image generation.
Compared to the operation on raw RGB pixels, the semantic map enables simpler user modification.
In this work, we specifically target at generating semantic maps given a label-set consisting of desired categories.
The proposed framework, SegVAE, synthesizes semantic maps in an iterative manner using conditional variational autoencoder.
Quantitative and qualitative experiments demonstrate that the proposed model can generate realistic and diverse semantic maps.
We also apply an off-the-shelf image-to-image translation model to generate realistic RGB images to better understand the quality of the synthesized semantic maps.
Furthermore, we showcase several real-world image-editing applications including object removal, object insertion, and object replacement.
\end{abstract}

% 

% text
\section{Introduction}
%\vspace{\secmargin}
%\vskip \secmargin
\vskip -1.2em
The recent success of deep generative models has made breakthroughs in a wide range of tasks such as image and video synthesis~\cite{goodfellow2014generative,karras2019style,kingma2013auto}. 
In addition to conventional generative models that aims to generate images from noise vectors sampled from prior distributions, conditional generative models have getting attention to handle various tasks including image-to-image translation (I2I)~\cite{isola2017image,DRIT,CycleGAN2017}, text-to-image synthesis~\cite{hong2018inferring,han2017stackgan}, and audio-to-video generation~\cite{lee2019dancing2music}, to name a few. 
One major goal of these conditional generation tasks is to enable flexible user control, image editing, and content creation.
Conditional generative models can greatly shorten the distance between professional creators and general users.

Among all forms of conditional context, semantic maps are recently getting attention.
Semantic maps can be used as an intermediate representation or directly as inputs.
As an intermediate representation, semantic maps serve as mediums that facilitate the original tasks such as image generation from text and scene graph~\cite{hong2018inferring,johnson2018image}.
As inputs, semantic maps can be translated to realistic images via I2I models~\cite{park2019SPADE}.
These I2I models enable editing on the semantic maps, which is easier and more flexible than operating on the RGB space.
However, in terms of image editing, users need to create semantic maps that are realistic in order to generate realistic RGB images.
It is crucial to provide users greater flexibility and less overhead on image editing.

In this work, we focus on generating semantic maps given a label-set.
The task is challenging for two reasons.
First, the shape of components in the semantic maps not only need to be realistic, but also have to be mutually compatible.
Second, the generation is inherently multimodal, that is, one label-set can correspond to multiple semantic maps.
To handle these issues, we propose \textbf{SegVAE}, a VAE-based framework that can generate semantic maps in an iterative manner.
For \textit{compatibility}, the proposed model performs generation at each iteration conditioned on the label-set as well as previously generated components.
For \textit{diversity}, the proposed method learns a shape prior distribution that can be randomly sampled during the inference stage.

We evaluate the proposed methods through extensive qualitative and quantitative experiments.
We conduct experiments on two diverse datasets, the CelebAMaskHQ~\cite{CelebAMask-HQ} and HumanParsing~\cite{liang2015human} datasets, to demonstrate the general effectiveness of the proposed framework.
We leverage Fr\'echet Inception Distance (FID)~\cite{fid} and conduct a user study to evaluate realism.
For diversity, we measure feature distances similar to the Learned Perceptual Image Patch Similarity (LPIPS)~\cite{zhang2018lpips} metric.
Furthermore, we demonstrate several real-world editing scenarios to showcase the superb controllability of the proposed method.
We also apply an I2I model, SPADE~\cite{park2019SPADE}, to synthesize realistic images based on the generated semantic maps to better visualize the quality of the proposed model.

We make the following contributions:
\begin{compactitem}[$\bullet$]
    \item We propose a new approach that can generate semantic maps from label-sets.
    Components in the generated semantic maps are mutually compatible and the overall semantic maps are realistic.
    SegVAE can also generate diverse results.
    %The proposed method can also generate diverse results.
    \item We validate that our method performs favorably against existing methods and baselines in terms of realism and diversity on the CelebAMask-HQ and HumanParsing datasets.
    \item We demonstrate several real-world image editing applications using the proposed method. Our model enables flexible editing and manipulation.
\charles{Our code and more results is available at \href{https://github.com/yccyenchicheng/SegVAE}{https://github.com/yccyenchicheng/SegVAE}.}

\end{compactitem}

\vskip \secmargin
\section{Related Work}
%\vspace{\secmargin}
\vskip \secmargin
\noindent\textbf{Generative models.}
Generative models aim to model a data distribution given a set of samples from that distribution.
The mainstream of generative models approximates the distribution through maximum likelihood.
There are two branches of the maximum likelihood method.
One stream of work explicitly models the data distribution.
PixelRNN~\cite{oord2016pixel} and PixelCNN~\cite{van2016conditional} are auto-regressive models that perform sequential generation where the next pixel value is predicted conditioned on all the previously generated pixel values.
Variational autoencoder~\cite{kingma2013auto} model the real data distribution by maximizing the lower bound of the data log-likelihood.
The other stream of work learns implicit data distribution.
Generative adversarial networks~\cite{goodfellow2014generative} model the data distribution by a two-player game between a generator and a discriminator.

Based on conventional generative models, conditional generative models synthesize
images based on various contexts. 
Conditional generative models can formulate a variety of topics in image editing and content creations, including super-resolution~\cite{ledig2017photo}, image-to-image translation~\cite{DRIT_plus,CycleGAN2017}, text-to-image generation~\cite{tseng2020retrievegan,han2017stackgan}, video generation~\cite{li2018video,p2pvg2019}, and music-to-dance translation~\cite{lee2019dancing2music}.

\noindent\textbf{Generative models with semantic maps.}
Semantic map is an important modality in generative modeling.
There are two major ways of using semantic maps.
First, semantic maps can be used as the conditional context for conditional image synthesis.
Image-to-image translation models can learn the mapping from semantic maps to realistic RGB images~\cite{isola2017image,pan2019video,park2019SPADE,wang2018pix2pixHD}.
Second, semantic maps can serve as an intermediate representation during the training of image synthesis conditioned on text or scene graph.~\cite{johnson2018image,li2019object,sun2019imagesf,talavera2019layout,tripathi2019heuristics}.
Using semantic layouts provides rough guidance of the location and appearance of objects and further facilitate the training.
In this work, we focus on generating semantic maps directly from a label-set.
The proposed model enables flexible user editing and the generated results can be further used for photorealistic image generation. 

\noindent\textbf{Image editing.}
Image editing is a process of altering images.
Traditional image editing mostly focuses on low-level changing like image enhancement, image retouching, and image colorization.
In this work, we target at generating and altering the content of images.
For generation, previous work can synthesize desired output given instructions like a set of categories~\cite{jyothi2019layoutvae,lee2019neural} or a sentence~\cite{tan2019text2scene}, \charles{or change the semantic information in a user-specified region of the target image~\cite{suzuki18}.}
Other stream of work can perform operations like object insertion~\cite{lee2018context,lin2018st}, object removal~\cite{yang2014automatic}, and object attribute editing~\cite{tseng2020art}.
The proposed method can achieve both generations: given specified label-sets, and editing including object insertion and removal.
%------------------------------------------------------------------------
\begin{figure*}[t!]
    \centering
    \includegraphics[width=\linewidth]{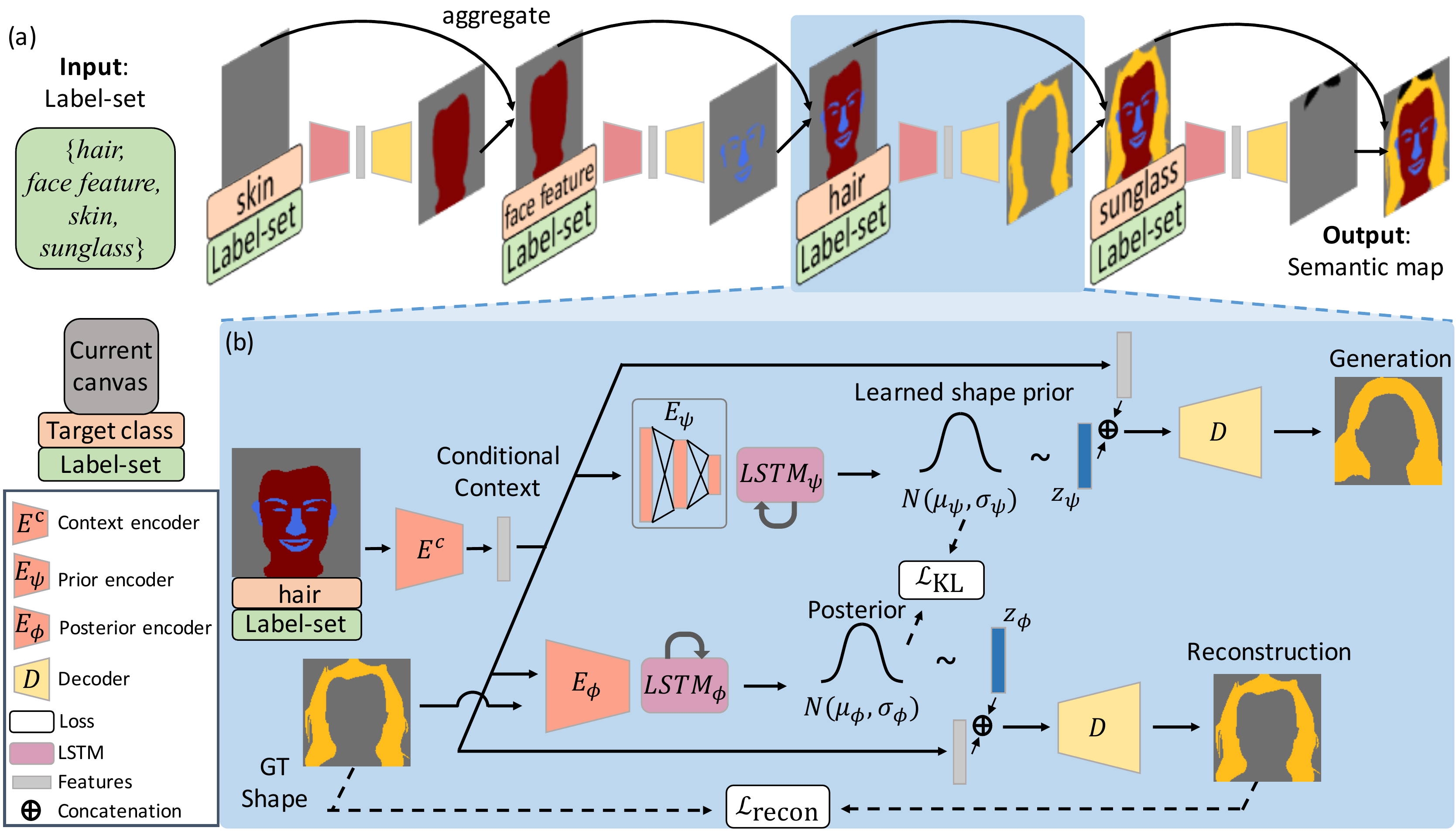}
    \caption{
    \textbf{Overview.}
    (a) Given a label-set as input, we iteratively predict the semantic map of each category starting from a blank canvas. we initialize the initial canvas as a blank semantic map $x_0$.
    The generation conditions on the embedding of (\textit{Current canvas, Target class, Label-set}).
    (b) In each iteration, the context encoder $E^{\mathrm{c}}$ encodes the input into the conditional context which is copied as the input to $E_{\phi}$, $E_{\psi}$, and $D$.
    During training, the posterior encoder $\{E_{\phi}, LSTM_{\phi}\}$ takes the ground-truth semantic map of the target class as additional input to output $N(\mu_{\phi}, \sigma_{\phi})$, which is then used to reconstruct the ground-truth semantic map.
    The prior encoder $\{E_{\psi}, LSTM_{\psi}\}$ encodes the conditional context to output $N(\mu_{\psi}, \sigma_{\psi})$, which enables the stochastic sampling during test time.
    }
    \vskip \figmargin
    \label{fig:overview}
\end{figure*}
%\vspace{\secmargin}
%\vskip \secmargin
\vskip \secmargin
\section{Semantic Maps Generation}
%\vspace{\secmargin}
%\vskip \secmargin
\vskip \secmargin
Our goal is to learn the mapping from label-sets $\mathcal{X}$ to semantic maps $\mathcal{Y} \subset \mathbb{R}^{H \times W \times C}$.
A label-set $\mathbf{x} = [x_1, x_2, ..., x_c] \in \mathcal{X}$ is a binary vector with $c$ classes, where $x_i \in \{0, 1\}$.
A semantic map $y \in \mathcal{Y}$ denotes a $c$ channels label map where $Y_{i, j, k} \in \{0, 1\}$ for class $k$, $k = 1, 2, ..., C$.
The proposed \textbf{SegVAE} consist of a context encoder $E_\mathrm{c}$, a posterior encoders $\{E_\phi, LSTM_\phi\}$, a learned prior encoders $\{E_\psi, LSTM_\psi\}$, and a decoder $D$.
The context encoder $E^\mathrm{c}$ aims to encode information including label-set, current canvas, and target label into a context for the following conditional generation.
The posterior network $\{E_\phi, LSTM_\phi\}$, consisting of a convolutional encoder and an LSTM, aims to conditionally encode input each iteration into an inference prior.
The learned prior network $\{E_\psi, LSTM_\psi\}$, consisting of a fully-connected encoder and an LSTM, targets at learning a conditional prior distribution to be sampled from during inference time.
The decoder $D$ then learns the mapping from both posterior and learned prior distribution to semantic maps.

In this section, we first introduce the background knowledge of conditional variational auto-encoder in \subsecref{cvae}.
Then we detail the proposed model in \subsecref{network}.
Finally, we provide implementation details in \subsecref{detail}.

\vskip \subsecmargin
\subsection{Conditional VAE}
\label{subsec:cvae}
%\vspace{\subsecmargin}
\vskip \subsecmargin

\noindent\textbf{Variational autoencoder.} 
A Variational Autoencoder (VAE)~\cite{kingma2013auto,rezende2014stochastic} attempts to explicitly model the real data distribution by maximizing the lower bound of the data log-likelihood.
VAE leverages a simple prior $p_\theta(\mathbf{z})$ (\eg, Gaussian) and a complex likelihood $p_\theta(\mathbf{x} | \mathbf{z})$ on latent variable $\mathbf{z}$ to maximize the data likelihood $p_\theta(\mathbf{x})$. 
An inference network $q_\phi(\mathbf{z}|\mathbf{x})$ is introduced to approximate the intractable latent posterior $p_\theta(\mathbf{z}|\mathbf{x})$. Here $\theta$ and $\phi$ denote the parameters of the generation and inference networks. We then jointly optimize over $\theta$ and $\phi$,
\begin{equation}
	\begin{split}
	\log p_\theta(\mathbf{x}) &= \log \int_{\mathbf{z}}p_\theta(\mathbf{x}|\mathbf{z})p(\mathbf{z}) \, d\mathbf{z}\\
	\hspace{-2em} & \geq \mathbb{E}_{q_\phi(\mathbf{z}|\mathbf{x})}\log p_\theta(\mathbf{x}|\mathbf{z}) - D_\mathrm{KL}(q_\phi(\mathbf{z}|\mathbf{x})||p(\mathbf{z})) \,.
    \end{split}
    \label{eq:vae}
    %\vspace{-0.6em}
\end{equation}
\vskip \eqmargin
With this inequality, the variational autoencoder aims to reconstruct data $\mathbf{x}$ with latent variable $\mathbf{z}$ sampled from the posterior $q_\phi(\mathbf{z}|\mathbf{x})$ while minimizing the KL-divergence between the prior $p(\mathbf{z})$ and posterior $q_\phi(\mathbf{z}|\mathbf{x})$.

\noindent\textbf{Conditional variational autoencoder.} A Conditional Variational Autoencoder (C-VAE)~\cite{sohn2015learning} is an extension of the VAE which condition on a prior information described by a variable or feature vector $\mathbf{c}$. The generation and inference network's output will base on the conditional variable $\mathbf{c}$, and the optimization over $\theta$ and $\phi$ becomes,
\begin{equation}
	\begin{split}
	\log p_\theta(\mathbf{x,c}) &= \log \int_{\mathbf{z}}p_\theta(\mathbf{x}|\mathbf{z,c})p(\mathbf{z}) \, d\mathbf{z}\\
	 \hspace{-2em} & \geq \mathbb{E}_{q_\phi(\mathbf{z}|\mathbf{x})}\log p_\theta(\mathbf{x}|\mathbf{z,c}) - D_\mathrm{KL}(q_\phi(\mathbf{z}|\mathbf{x,c})||p(\mathbf{z|c})) \,.
    \end{split}
    \label{eq:cvae}
\end{equation}
\vskip \eqmargin

\vskip \subsecmargin
\subsection{Iterative Generation with Learned Prior}
\label{subsec:network}
%\vspace{\subsecmargin}
\vskip \subsecmargin
A label-set consists of various amount of categories.
To handle the dynamic amount of categories as well as capture dependency among categories, we leverage Long-Short Term Memory (LSTM)~\cite{hochreiter1997long} as our recurrent backbone.
However, how to perform stochastic generation and model data variations remain a challenge.
The complexity is twofold.
First, there are various possible combinations of categories to construct a label-set.
For example, $y_1=\{\textit{t-shirt}, short, shoe\}$ and $y_2=\{sunglasses, dress, bag\}$ represent two completely different outfit style.
Second, in addition to label-set-level variations, category-level variations also need to be captured.
For instance, each \textit{bag} and \textit{dress} have diverse possible shapes.
Moreover, the label-set-level variations and the category-level variations are not independent.
For example, different choices of \textit{t-shirt} will affect the choices of \textit{dress} considering compatibility.

To handle these issues, we leverage C-VAE as our generative model.
However, the usage of $\mathcal{N}(\mathbf{0},\mathbf{I})$ which is conventionally used in VAE ignores dependencies between iterations since the priors are drawn independently at each iteration.
Therefore, we adopt the similar idea from~\cite{denton2018stochastic} to use a parameterized network $\psi$ other than $\mathcal{N}(\mathbf{0},\mathbf{I})$ for inferring the shape prior distribution.
In order to learn the conditional prior, we first define our conditional context at iteration $t$ as:
\begin{equation}
    %c_t = E^c(y, x_{1:t-1}),
    c_t = E^c(y_t, y, x_{1:t-1}),
\end{equation}
where $y_t$ is the current category to be generated, $y$ represents the given label set, and $x_k$ denote the semantic map at iteration $k$.
The learned prior can thus be modeled with $\{E_\psi, LSTM_\psi\}$.
At iteration $t$, 

\begin{equation}
\begin{aligned}
       h_t & = E_\psi(c_t)\\
      %\mu^{t}_\psi, \sigma^{t}_\psi & = \mathrm{LSTM}_\psi(h_t, y_t)\\
      \mu^{t}_\psi, \sigma^{t}_\psi & = \mathrm{LSTM}_\psi(h_t)\\
      z_t & \sim \mathcal{N}(\mu^{t}_\psi, \sigma^{t}_\psi).
\end{aligned}
\end{equation}
The learned prior is trained with the help of the posterior network $\{E_\phi, LSTM_\phi\}$ and the decoder $D$.
At iteration $t$, given $x_t$, the semantic map of current category, we perform reconstruction by
\begin{equation}
\begin{aligned}
      %h_t & = E_\phi(c_t)\\
      %\mu^{t}_\phi, \sigma^{t}_\phi & = \mathrm{LSTM}_\phi(h_t, y_t)\\
      h_t & = E_\phi(c_t, x_t)\\
      \mu^{t}_\phi, \sigma^{t}_\phi & = \mathrm{LSTM}_\phi(h_t)\\
      z_t & \sim \mathcal{N}(\mu^{t}_\phi, \sigma^{t}_\phi)\\
      \hat{x}_t & = D(z_t, c_t).
\end{aligned}
\end{equation}
Therefore, the model can be trained with a reconstruction loss and a KL loss at iteration $t$:
\begin{equation}
    \begin{aligned}
    L_\mathrm{recon}^t & = \lVert \hat{x}_t - x_t\lVert_{1}\\
    L_\mathrm{KL}^t & = -D_\mathrm{KL}(\mathcal{N}(\mu^{t}_\phi, \sigma^{t}_\phi) || \mathcal{N}(\mu^{t}_\psi, \sigma^{t}_\psi)).
    \end{aligned}
\end{equation}
We express the final objective as:
\begin{equation}
    L = \mathbb{E}_{x,y}\big[ \sum_{i=1}^{|y|} \lambda_\mathrm{recon}L_\mathrm{recon}^i + \lambda_\mathrm{KL}L_\mathrm{KL}^i \big],
\end{equation}
where the hyper-parameters $\lambda$s control the importance of both term.

In the inference time, given a label-set as input, we initialize an empty label map as $x_0$.
We then generate $x_k$ autoregressively by setting the inputs for $p_\psi$ with the generated shapes $\hat{x}_{1:t-1}$.

\vskip \subsecmargin
\subsection{Implementation Details}
\label{subsec:detail}
%\vspace{\subsecmargin}
\vskip \subsecmargin
We implement our model in PyTorch~\cite{pytorch}. 
For all experiments, we use the resolution of $128 \times 128$ for the input image and semantic map.
For the context encoder $E^{c}$, we use two multilayer perceptrons to produce the embeddings of the label-set and target label, and fuse them with the current canvas.
Then a six convolution layers will encode the embeddings and the current canvas into the conditional context.
For the $E_{\psi}$, we apply a fully connected layers with LSTM followed by two multilayer perceptrons to output the mean and log variance.
The $E_{\phi}$ is similar to $E_{\psi}$ by replacing the fully connected layers with the convolution layers to encoder the ground truth semantic map.
We use the latent code size of $z_{\psi}, z_{\phi} \in \mathbb{R}^{384}$ for all experiments.
Finally, the $D$ consists of five fractionally strided convolution layers.
We apply the instance norm and spectral norm for all the architectures.
For more details of the network architecture, please refer to the supplementary material.

For training, we use the Adam optimizer~\cite{adam} with a learning rate of $5e^{-5}$, a batch size of $24$, and $(\beta_1, \beta_{2})=(0.5, 0.999)$.
For the HumanParsing dataset, we use $\lambda_{\mathrm{recon}}=1$,  $\lambda_{\mathrm{KL}}=1e^{-4}$ and $\lambda_{\mathrm{recon}}=1$, $\lambda_{\mathrm{KL}}=1e{-7}$ for the CelebAMask-HQ dataset.
For the order of the iterative generation with learned prior, we use $\{$\textit{face, hair, left arm, right arm, left leg, right leg, upper clothes, dress, skirt, pants, left shoe, right shoe, hat, sunglasses, belt, scarf, bag}$\}$ for the HumanParsing dataset. 
In addition, $\{$\textit{skin, neck, hair, left eyebrow, right eyebrow, left ear, right ear, left eye, right eye, nose, lower lip, upper lip, mouth, hat, cloth, eyeglass, earrings, necklace}$\}$ for the CelebAMask-HQ dataset.
%
%------------------------------------------------------------------------
%\vspace{\secmargin}
\vskip \secmargin
\section{Experimental Results}
%\vspace{\secmargin}
\vskip \secmargin
\begin{figure}[t]
    \centering
    \includegraphics[width=\linewidth]{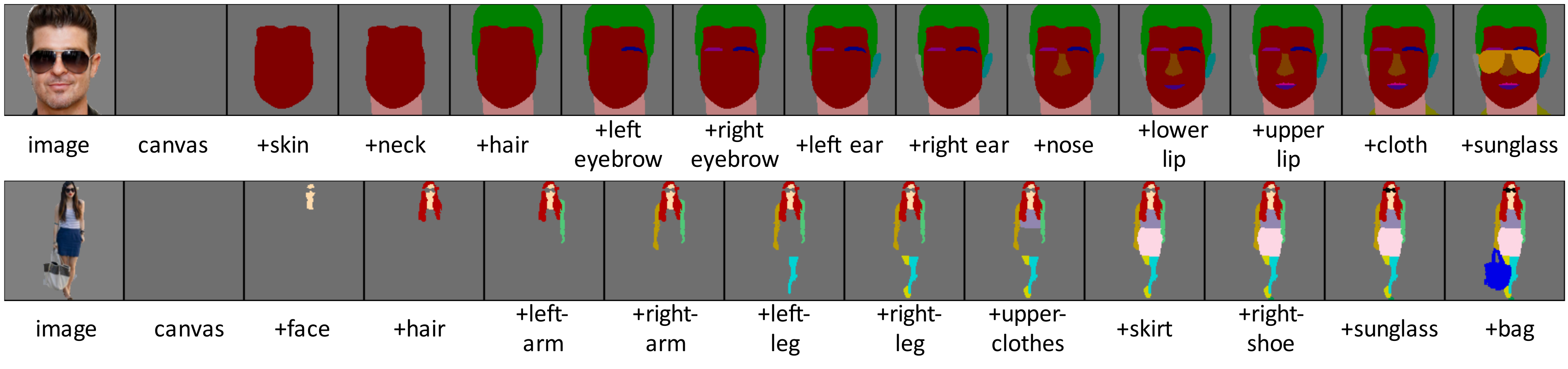}
    \vspace{-1.5em}
    \caption{\textbf{Training data example.}
    Starting from an empty canvas, we generate semantic map in an iterative manner, where each generation conditions on all previous generated semantic maps. We show the visualization of the training data for the proposed method. Starting from an empty canvas, we generate semantic map in an iterative manner, where each generation conditions on all previous generated semantic maps.
    } 
    \label{fig:comp_baselines} 
    \vskip \figmargin
\end{figure}
% table
\begin{table*}[t]
\caption{\tb{Realism and Diversity.} We use the FID and a feature space distance metric to measure the realism and diversity ($\rpm$ indicates the $95 \%$ C.I.).}
\centering
\begin{tabular}{l  cc cc} 
    \toprule
     & \multicolumn{2}{c}{HumanParsing} & \multicolumn{2}{c}{CelebAMask-HQ} \\
    \cmidrule(r){2-3} \cmidrule(r){4-5}
    {Method} & {FID $\shortdownarrow$ } & {Diversity $\shortuparrow$ } & {FID $\shortdownarrow$ } & {Diversity $\shortuparrow$ } \\
    \midrule
    $\text{C-GAN}$  & $171.1293^{\rpm .3359}$ & N/A & $76.0115^{\rpm .1981}$ & N/A \\
    $\text{C-VAE}_{\text{sep}}$  & $85.0505^{\rpm .3052}$ & $.1781^{\rpm .040}$  & $39.7445^{\rpm .2254}$ & ${.1566^{\rpm .033}}$ \\
    $\text{C-VAE}_{\text{global}}$ & $83.8214^{\rpm .6747}$ & $.1730^{\rpm .034}$  & $36.2903^{\rpm .2084}$ & $\mathbf{.1582^{\rpm.031}}$ \\
    sg2im-orig~\cite{johnson2018image} & $207.0786^{\rpm .3324}$ & N/A & $208.8142^{\rpm .1876}$ & N/A \\
    sg2im~\cite{johnson2018image}  & $56.7421^{\rpm .2206}$ & \underline{$.2064$}$^{\rpm .050}$  & $34.7316^{\rpm .3071}$ & $.1476^{\rpm .045}$ \\
    \midrule
    Ours w/o LSTM  & $50.8830^{\rpm .2374}$ & $.2024^{\rpm .045}$  & $34.5073^{\rpm .2156}$ & $.1535^{\rpm .034}$ \\
    Ours w/o Learned Prior   & \underline{$44.6217$}$^{\rpm .2881}$ & $.1625^{\rpm .054}$  & \underline{$33.8060$}$^{\rpm .3167}$ & $.1492^{\rpm .038}$ \\
    %Ours  & - & - & - & - & - & - \\
    Ours & $\mathbf{39.6496^{\rpm .3543}}$ & $\mathbf{.2072^{\rpm .053}}$ & $\mathbf{28.8221}^{\rpm .2732}$ & \underline{$.1575$}$^{\rpm .043}$ \\
    \midrule
    %Ours (Recon.) & - & 37.1793 & - & - & 26.1576 & - \\ 
    GT  & $33.1562^{\rpm .3801}$ & $.2098^{\rpm .050}$  & $22.5981^{\rpm .0870}$ & $.1603^{\rpm .045}$ \\
    \bottomrule
\end{tabular}
%\vspace{\tabmargin}
\vskip \tabmargin
\label{tab:quan_fid_div}
\end{table*}

\paragraph{Datasets.} We perform the evaluation on two datasets:
\begin{compactitem}[$\bullet$]
    \item\textbf{HumanParsing.} 
The HumanParsing dataset~\cite{liang2015human}, extended from~\cite{liang2015atr}, contains $\sim 17,000$ of street fashion images and 18 classes of semantic maps with pixel-annotations.
It is composed of diverse appearances and multiple combinations of fashion items and human parts.
We first clean the dataset by inspecting the aspect ratio of the ground truth semantic maps and remove those over $\rpm 1$ standard deviation.
For each example, we compute the bounding box of each class on the semantic map and crop them accordingly.% for training.
\item\textbf{CelebAMask-HQ.}
CelebAMask-HQ~\cite{CelebAMask-HQ} is a large-scale face image dataset which includes $30,000$ high-resolution paired face images and semantic maps. The semantic maps contain 19 classes of facial attributes and accessories.
The various shapes of facial attributes and accessories make it suitable for testing.% the proposed methods.
\end{compactitem}

We split the data into $80\%, 10\%, 10\%$ for training, validation and testing. \charles{\noindent For more quantitative and qualitative results, please refer to the supplementary files.}%for all experiments.

\paragraph{Evaluated Method.}
We evaluate the following algorithms.
\begin{compactitem}[$\bullet$]
    \charles{
    \item $\mathbf{C\text{-}{GAN}}$.
    We implement a conditional GAN which takes a label-set as input and generates the corresponding images.
    %We compare this method by computing the FID on two datasets in \tabref{quan_cgan}.
    This method has to handle the class-dependency, compatibility between shapes, and the image quality at the same time, which imposes a great burden on a single model.
    }
    \item $\mathbf{C\text{-}VAE_{sep}}$.
    This baseline generates semantic maps for each category independently.
    The generated semantic maps are then aggregated to the final output.

    \item $\mathbf{C\text{-}VAE_{global}}$
    This baseline takes global context into consideration while generating semantic maps for each category.
    %
    % We encode the label-set into a global feature, which serves as the conditional context for the C-VAEs generation.
    We encode the label-set into a global feature, which serves as the conditional context for the generation.
    With reference to the global feature, the generated shapes will be more compatible.
    
    \item \textbf{sg2im~\cite{johnson2018image}.}
    Sg2im is a conditional generative model that takes a scene graph as input and output the corresponding image.
    To facilitate the learning, a semantic map is first predicted from the object embedding during the optimization process.
    We compare the predicted semantic maps from sg2im given a label-set as input, which can be seen as the simplest form of a scene graph.
    We use the official implementation provided by \cite{johnson2018image} for training.
    For a fair comparison, we provide ground truth bounding boxes for sg2im when generating the masks.
    We report both the metrics from the images translated from its predicted semantic maps using SPADE~\cite{park2019SPADE} and the images generated by sg2im (denoted by ``sg2im'' and ``sg2im-orig'' respectively in \tabref{quan_fid_div}).
\end{compactitem}

\begin{table*}[t]
\small
\caption{\tb{Compatibility and reconstructability.} 
We train a shape predictor to measure the compatibility error (abbreviated as Compat. error) over the generated shapes. % for all methods.
We adopt an auto-encoder to measure the quality of our generated results by computing the reconstruction error (denoted as Recon. error).}
\centering
\begin{tabular}{l  cc cc} 
    \toprule
     & \multicolumn{2}{c}{HumanParsing} & \multicolumn{2}{c}{CelebAMask-HQ} \\
    \cmidrule(r){2-3} \cmidrule(r){4-5}
    {Method} & {Compat. error~$\shortdownarrow$} & {Recon. error~$\shortdownarrow$} & {Compat. error~$\shortdownarrow$} & {Recon. error~$\shortdownarrow$} \\
    \midrule
     $\text{C-VAE}_{\text{sep}}$ & $.7823^{\rpm .0161}$ & $.6857^{\rpm .010}$ & $.1029^{\rpm .0017}$ & $.1165^{\rpm.003}$ \\
     $\text{C-VAE}_{\text{global}}$ & $.7345^{\rpm .0141}$ & $.6186^{\rpm .018}$ & $.1015^{\rpm .0040}$ & $.1142^{\rpm.003}$ \\
    sg2im~\cite{johnson2018image}     & $ .6983^{\rpm .0176}$ & $\mathbf{.5434^{\rpm .012}}$ & $.0844^{\rpm .0020}$ & $.1334^{\rpm .003}$ \\
    Ours & $\mathbf{.6174^{\rpm .0147}}$ & $.5663^{\rpm .011}$ & $\mathbf{.0754^{\rpm .0013}}$ & $\mathbf{.0840^{\rpm.001}}$ \\
    \bottomrule
\end{tabular}
%\vspace{\tabmargin}
\vskip \tabmargin
\label{tab:quan_compat_recon}
\end{table*}

%\vspace{\subsecmargin}
\vskip \subsecmargin
\subsection{Quantitative Evaluation}
\label{subsec:quantitative}
%\vspace{\subsecmargin}
\vskip \subsecmargin

\noindent\textbf{Visual quality.}
We evaluate the realism of the generated semantic maps using the Fr\'echet Inception Distance (FID)~\cite{fid} metric.
We first generate a semantic map given a label-set, then we use an off-the-shelf I2I model~\cite{park2019SPADE} to output the translated image.
We compute FID with the official implementation from \cite{fid} on all compared methods to measure the realism of the generated semantic maps.
\tabref{quan_fid_div} shows that the proposed model performs favorably against the baselines and the existing method.

%\vspace{\paramargin}
\vskip \paramargin
\noindent\textbf{Diversity.}
We measure the diversity by computing distance between the generated semantic maps using the distance metric similar to the LPIPS~\cite{zhang2018lpips} metric.
However, there are no general feature extractors for semantic maps.
Therefore, we trained an auto-encoder on the ground truth semantic maps.
We use the intermediate representation as the extracted features for semantic maps.
We can then measure the feature distance between two semantic maps.
We measure the diversity between 3000 pairs of generated semantic maps by sampling from the test-set.
\tabref{quan_fid_div} shows that the proposed method generates diverse semantic maps without sacrificing visual quality.

%\vspace{\paramargin}
\vskip \paramargin
\noindent\textbf{Compatibility and reconstruction error.}
To evaluate the compatibility of a generated semantic maps, we design a metric measure the performance of all compared methods.
A shape is compatible if it can be easily inferred from all the other class's semantic maps.
We measure this quantitatively by training a shape predictor which takes a semantic map and a target label as input, and outputs the shape of the target class.
The training pair of the data is created by excluding one of the class in the ground truth semantic map as the prediction target, and the remaining classes' semantic maps along with the class of the excluded target form the input.
Then we use this shape predictor to compute the compatibility error for all the compared methods.
Meanwhile, we also train an auto-encoder on the ground-truth semantic maps to measure if one generated result is close to the real data distribution.
Given a generated semantic map as input, we calculate the reconstruction error between the input and the reconstructed result output by the auto-encoder.
\tabref{quan_compat_recon} shows that the proposed method generates compatible and reasonable semantic maps.

\begin{table}[t]
\caption{\textbf{Prediction order analysis}. (Upper) On the HumanParsing dataset, order--1 and order--2 perform similarly since both \textit{Body, Clothes} provide great contexts for the model. In contrast, order--3 degrades the generation results and the diversity since \textit{Accessories} offers limited contexts.
(Lower) On the CelebAMask-HQ dataset, \textit{Face} is crucial for serving as the initial canvas for the subsequent generation. Therefore order--1 outperforms order--2 and order--3. in FID. Similarly, order--3 largely constrains the possible generation for the remaining class and causes lower diversity.
}
\centering
\small
\begin{tabular}{l  cc} 
    \toprule
     %& \multicolumn{2}{c}{HumanParsing} \\
    %\cmidrule(r){2-3}
    {Order}  (HumanParsing) & {FID}$\shortdownarrow$ & {Diversity}$\shortuparrow$ \\
    \midrule
    $\text{1}~\textit{Body} \rightarrow \textit{Clothes} \rightarrow \textit{Accessories}~\text{(Ours)}$ & $\mathbf{39.6496^{\rpm .3543}}$ & $\mathbf{.2072^{\rpm .053}}$ \\
    $\text{2}~\textit{Clothes} \rightarrow \textit{Body} \rightarrow \textit{Accessories}$ & $39.9008^{\rpm .5263}$ & $.2062^{\rpm .0494}$  \\
    $\text{3}~\textit{Accessories} \rightarrow \textit{Body} \rightarrow \textit{Clothes}$ & $40.2909^{\rpm .2195}$ & $.2043^{\rpm .0521}$ \\
    \bottomrule
\end{tabular}
\begin{tabular}{l  cc} 
    \toprule
    % & \multicolumn{2}{c}{CelebAMask-HQ} \\
    %\cmidrule(r){2-3}
    {Order} (CelebAMask-HQ) & {FID}$\shortdownarrow$ & {Diversity}$\shortuparrow$ \\
    \midrule
    $\text{1}~\textit{Face} \rightarrow \textit{Face features} \rightarrow \textit{Accessories}~\text{(Ours)}$ & $\mathbf{28.8221^{\rpm .2732}}$ & $\mathbf{.1575^{\rpm .043}}$ \\
    $\text{2}~\textit{Face features} \rightarrow \textit{Face} \rightarrow \textit{Accessories}$ & $30.6547^{\rpm .1267}$ & $.1517^{\rpm .0376}$ \\
    $\text{3}~\textit{Accessories} \rightarrow \textit{Face} \rightarrow \textit{Face features}$ & $32.0325^{\rpm .1294}$ & $.1489^{\rpm .0363}$ \\
    \bottomrule
\end{tabular}
\vskip \tabmargin
\label{tab:quan_order}
\end{table}

%\begin{figure}[h]
\begin{figure}[t]
    \centering
    \includegraphics[width=0.8\linewidth]{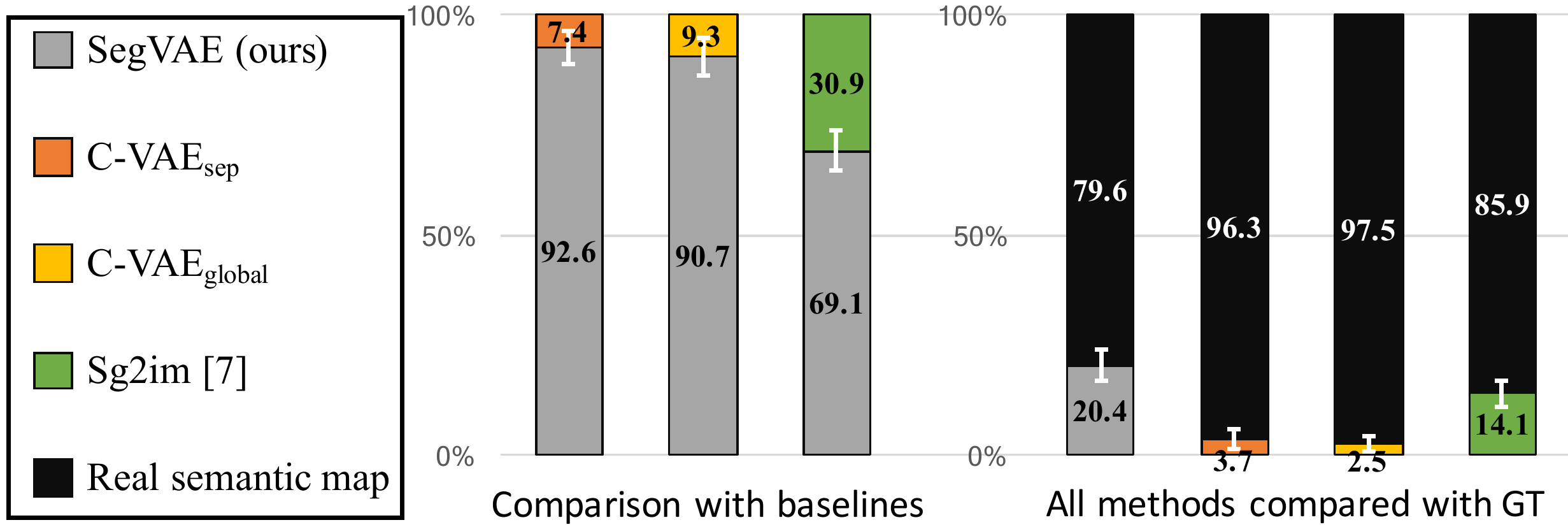}
    \caption{\textbf{User study.}
    We conduct a user study to evaluate the realism of the generated semantic maps.
    We have two sets of comparison: the comparison with other algorithms, and all methods to real maps.
    Results showed that users favored SegVAE (gray bar) against all other compared methods in both settings.
    }
    \label{fig:userstudy}
    \vskip \figmargin
\end{figure}

% "User study. We conduct a user study to evaluate the realism of the generated semantic maps. We have two sets of comparison: the comparison with other algorithms, and all methods to the ground truth. The charts showed that the users favored Ours (the gray bar) against all other compared methods in both settings."
\begin{figure}[t]
    \centering
    \includegraphics[width=\linewidth]{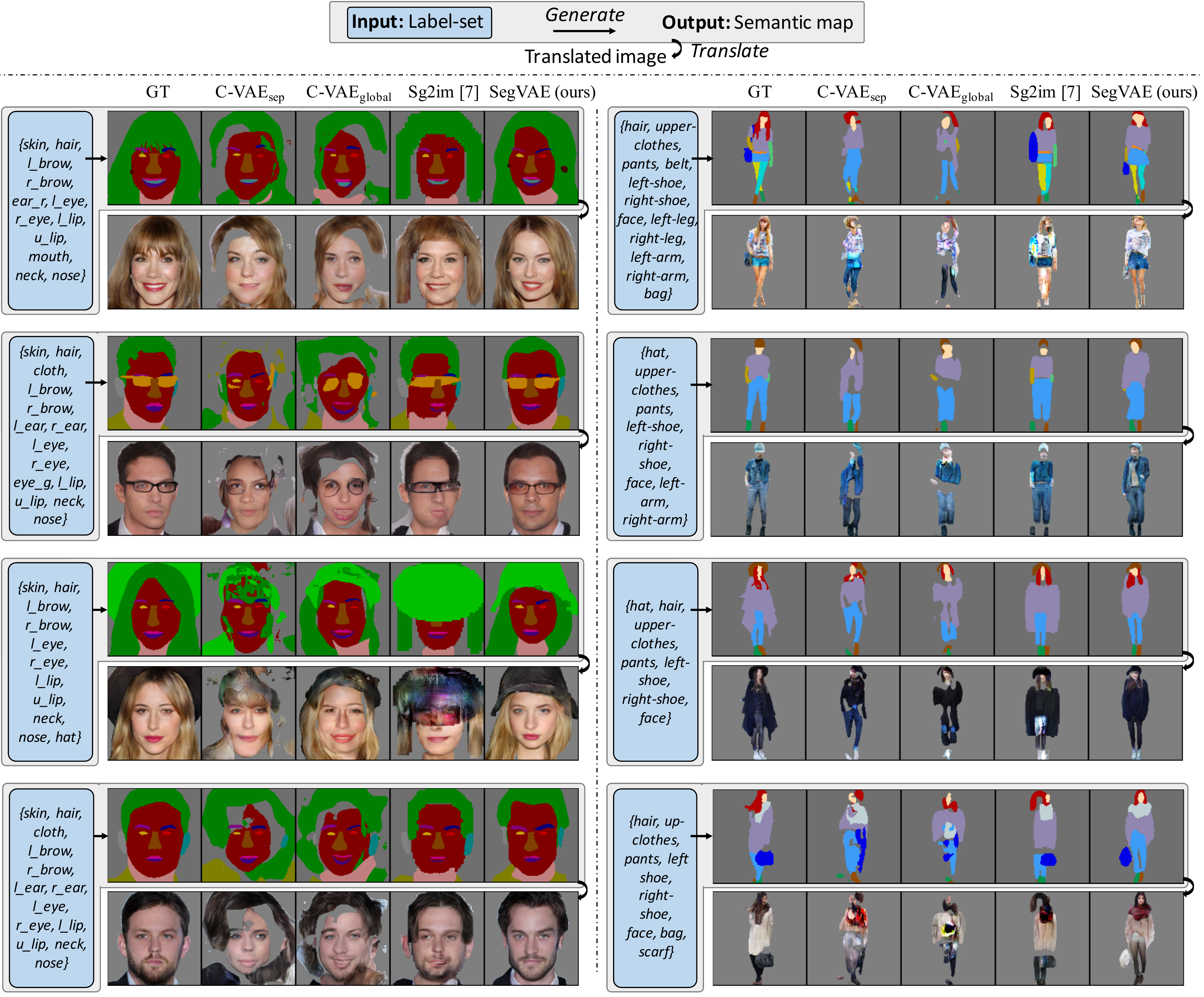}
    \vspace{-1.2em}
    \caption{\textbf{Qualitative comparison.} 
    We present the generated semantic maps given label-sets on the CelebAMask-HQ (left) and the HumanParsing (right) datasets.
    The proposed model generates images with better visual quality compared to other methods.
    We also present the translated realistic images via SPADE.
    Please refer to \figref{multi_modal} for the color mapping for each category. } 
    \label{fig:comp_baselines} 
    \vskip \figmargin
\end{figure}
%\input{tex/fig_multi}

%\vspace{\paramargin}
\vskip \paramargin
\noindent\textbf{Prediction order analysis.} 
We analyze how the prediction order will affect the performance of the proposed model at Tables~\ref{tab:quan_order}.
First, we organize the classes in each dataset into three major categories for two datasets where the class in each category has similar properties.
Then, we train the proposed methods with different permutations of these orders. 
{\flushleft \textit{HumanParsing.}}
We organize the classes into \textit{Body, Clothes, Accessories}. The classes in each category are
        \begin{itemize}[$\bullet$]
            \item \textit{Body}: $\{$\textit{face, hair, left arm, right arm, left leg, right leg}$\}$.
            \item \textit{Clothes}: $\{$\textit{upper clothes, dress, skirt, pants, left shoe, right shoe}$\}$.
            \item \textit{Accessories}: $\{$\textit{hat, sunglasses, belt, scarf, bag}$\}$.
        \end{itemize}
    Three orders are as follows. Order--1: \textit{Body, Clothes, Accessories} (Ours), order--2: \textit{Clothes, Body, Accessories}, and order--3: \textit{Accessories, Body, Clothes}.
{\flushleft \textit{CelebAMask-HQ.}} We organize the classes into \textit{Face, Face features, Accessories}. The classes in each category are
        \begin{itemize}[$\bullet$]
            \item \textit{Face}: $\{$\textit{skin, neck, hair}$\}$.
            \item \textit{Face features}: $\{$\textit{left eyebrow, right eyebrow, left ear, right ear, left eye, right eye, nose, lower lip, upper lip, mouth}$\}$.
            \item \textit{Accessories}: $\{$\textit{hat, cloth, eyeglass, earrings, necklace}$\}$.
        \end{itemize}
    We experiment on three prediction orders. Order--1: \textit{Face, Face features, Accessories} (Ours), order--2: \textit{Face features, Face, Accessories}, and order--3: \textit{Accessories, Face, Face features}.

Table~\ref{tab:quan_order} shows how the prediction order affects the performance of realism and diversity.
For the HumanParsing dataset, order--1 and order--2 perform similarly on FID and diversity since both \textit{Body} and \textit{Clothes} provide great contexts for the model.
While in order--3, \textit{Accessories} do not deliver good information as the other two categories and will constrain the possible shapes and locations when generating the semantic maps of \textit{Body} and \textit{Clothes}.
This results in a degradation in FID and diversity.
For the CelebAMask-HQ dataset, the \textit{Face} category is essential.
Acting as a canvas, it ensures the generation of the subsequent class is located in the semantically meaningful position, and gives the most degree of freedom when generating compatible shapes compared to other orders.

%\vspace{\paramargin}
\vskip \paramargin
\noindent\textbf{Ablation studies.}
We analyze the contribution of each component of our method with ablation studies. 

\begin{compactitem}[$\bullet$]
    \item \textbf{Ours w/o LSTM.}
    This model omits the LSTM when outputting the distribution of the learned prior and the posterior. 
    \item \textbf{Ours w/o Learned Shape Prior.}
    This version adopts a fixed Gaussian distribution for the shape prior. The proposed method with fixed prior.
\end{compactitem}
The results in \tabref{quan_fid_div} show the necessity of the proposed model's architecture design. LSTM serves as a crucial role to handle the dynamic among the classes as well as capture the dependency among them. On both datasets, the model yields significant improvements over FID in contrast to the model with fixed prior. On the other hand, adopting the learned shape prior mainly contributes to the better diversity of the generated outputs. Besides, realism also gains improvement on both datasets.
%

%\vspace{\paramargin}
\vskip \paramargin
\noindent\textbf{User study.}
%\label{subsec:user}
To better explicitly evaluate the visual quality of the generated semantic maps without relying on the off-the-shelf image-to-image translation model, we conduct a user study.
We perform two sets of comparisons.
First, we compare the proposed method to baselines.
Second, we compare all methods to ground-truth semantic maps.
In each test, users are given two semantic maps generated from different methods and asked ``which semantic map looks more realistic?''.
We ask each user to compare 20 pairs and collect results from a total of 60 subjects.
\figref{userstudy} presents the results of our user study.
\vskip -1.2em

\begin{figure}[t!]
    \centering
    \includegraphics[width=\linewidth]{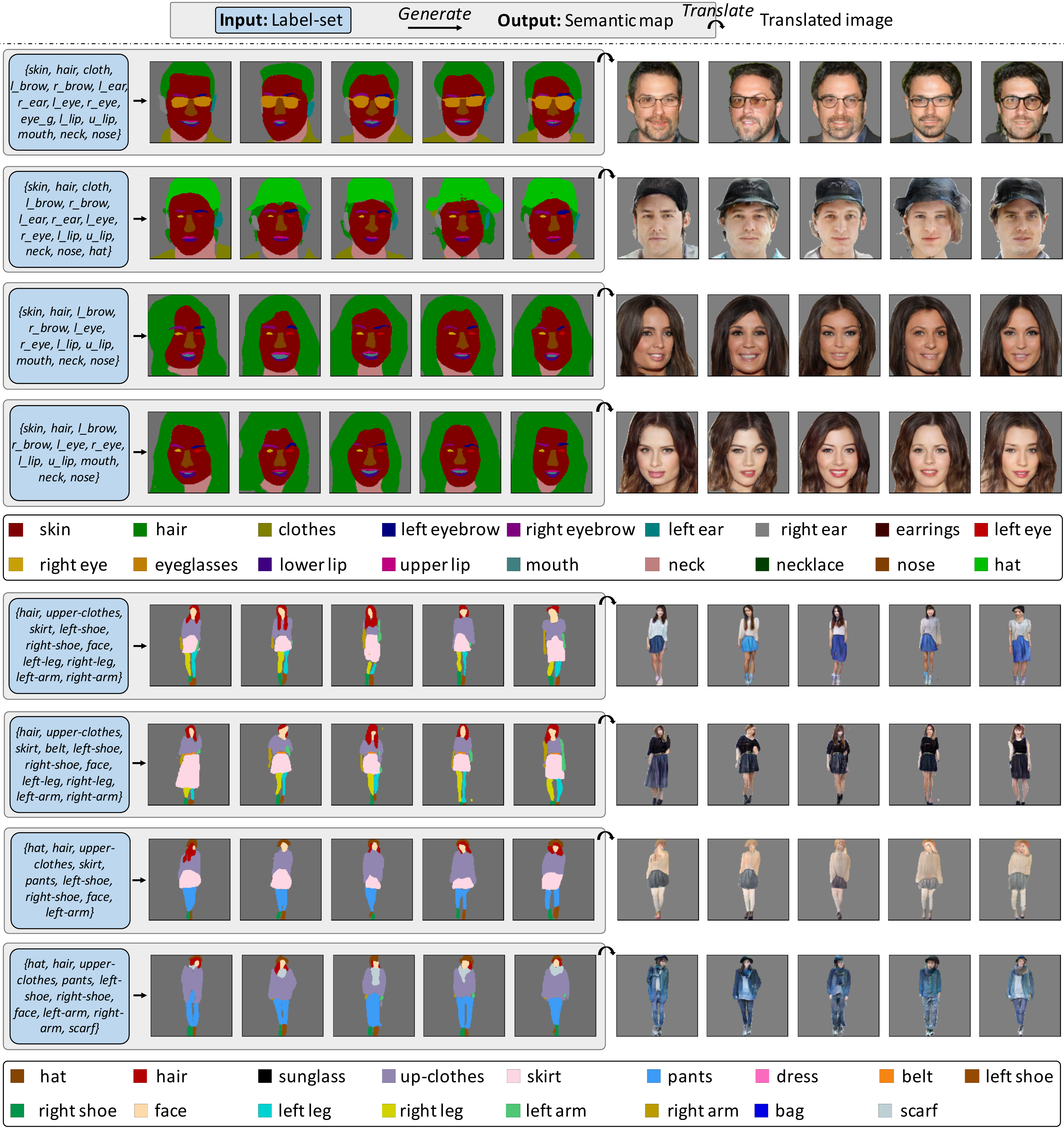}
    
    \caption{\textbf{Multi-modality.}
    We demonstrate the ability of SegVAE to generate diverse results given a label-set on both datasets.
    % We demonstrate the ability of the proposed model to generate diverse results given a label-set on the CelebAMask-HQ (top) and HumanParsing (bottom) dataset.
    }
    \label{fig:multi_modal} 
    \vspace{\figmargin}
\end{figure}

% figures
\begin{figure}[t!]
    \centering
    \includegraphics[width=\linewidth]{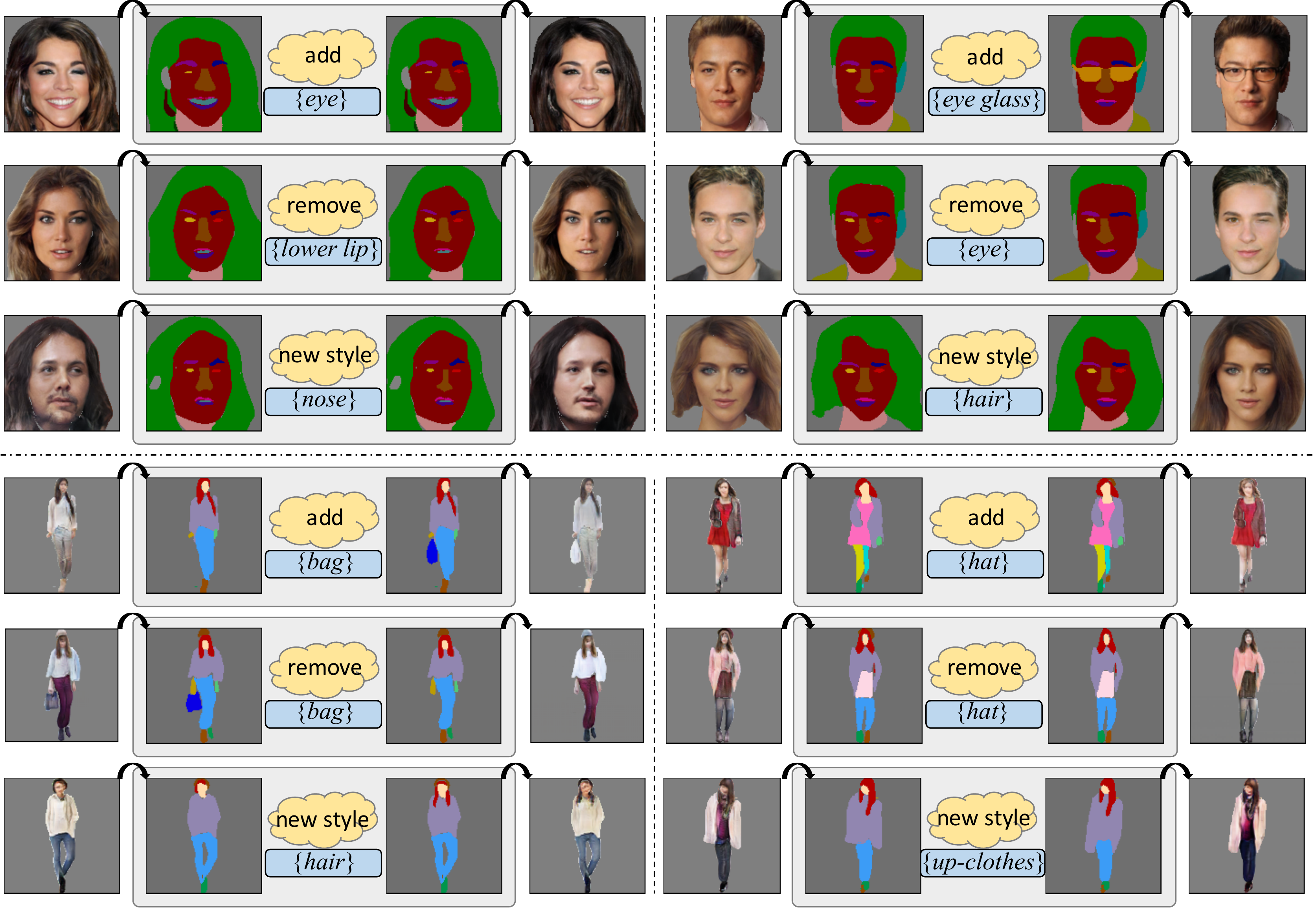}
 
    \caption{\textbf{Editing.}
    We present three real-world image editing applications: \textit{add}, \textit{remove}, and \textit{new style}.
    We show results of three operations on both datasets.
    % We show results of three operations on both CelebAMask-HQ and HumanParsing datasets.
    } 
    \label{fig:seg_editing} 
    \vskip \figmargin
\end{figure}

%\vspace{\subsecmargin}
\vskip \subsecmargin
\subsection{Qualitative Evaluation}
\label{subsec:qual_gen_multi}
%\vspace{\subsecmargin}
\vskip \subsecmargin
%\vskip -1em

%\vspace{\paramargin}
\vskip \paramargin

\noindent\textbf{Visual quality.} 
We compare the visual quality of the generated semantic maps from the proposed method and the baseline methods.
As shown in \figref{comp_baselines}, the proposed model can generate semantic maps that are more realistic than others.
The $\mathbf{C\text{-}VAE_{sep}}$ model generates each category independently.
The generation of each semantic map is not only unaware of the shape of other categories, but even unknowing of what other categories are in this label-set.
Although the individual semantic maps are reasonable, they are not mutually compatible after combination.
The $\mathbf{C\text{-}VAE_{global}}$ model performs slightly better with taking the label-set information as an additional input.
However, the generation process still disregards the appearance of each other.
Finally, the proposed SegVAE outperforms sg2im from several perspectives.
First, the iterative generation allows us to use teacher-forcing in training, which helps capture the dependency among components.
Second, we can better capture the appearance distribution with the help of VAE.
Finally, the learned latent space enables random sampling during inference time, which supports generating diverse outputs.

%\vspace{\paramargin}
\vskip \paramargin
\noindent\textbf{Multi-modality.}
We demonstrate the capability of the proposed method to generate diverse outputs in \figref{multi_modal}.
SegVAE can generate diverse appearances of objects given the same label-set.
For example, the eyeglasses (the first row of CelebAMask-HQ), the hat (the second row of CelebAMask-HQ), the skirt (the second row of HumanParsing), and the scarf (the fourth row of HumanParsing).
The generated semantic maps are mutually compatible thanks to the iterative conditional generation process.
In the fourth row of CelebAMask-HQ, the hair and face change jointly to different orientations.
In the first row of HumanParsing, both left- and right-leg changes jointly to different poses.
\vskip \subsecmargin

%\vspace{\subsecmargin}
%\vskip \subsecmargin
\vskip \subsecmargin
\subsection{Editing}
\label{subsec:qual_edit}
%\vspace{\subsecmargin}
\vskip \subsecmargin
The proposed method enables flexible user-control for image editing.
We demonstrate three real-world applications, including \textit{remove}, \textit{add}, and \textit{new style}.
For all applications, we first obtain the semantic maps from the input RGB images.
Then we perform editing on the semantic maps with SegVAE.
Finally, we translate the edited semantic maps back to RGB images with an I2I model.
We show the results of image editing in \figref{seg_editing}.
For \textit{Remove}, we remove a category from the label-set.
For \textit{Add}, we add an additional category to the existing label-set to perform object insertion.
For \textit{New style}, we could alter the style of the targeted category due to the diverse outputs of the purposed method.
More results are available in the supplementary materials.
%\vspace{-1.2em}
%\vskip -1.0em
%------------------------------------------------------------------------
%\vspace{\secmargin}
\vskip \secmargin
%\vskip -2.0em
\section{Conclusion}
%\vspace{\secmargin}
\vskip \secmargin
%\vskip -1.0em

In this paper, we present a novel VAE-based framework with iterative generation process for label-set to semantic maps generation. To generate realistic appearance and mutually compatible shapes among the classes, we propose to generate the target shape in an autoregressive fashion by conditioning on the previously generated components. To handle the diverse output, we adopt a parameterized network for learning a shape prior distribution. Qualitative and quantitative results show that the generated semantic maps are realistic and diverse. We also apply an image-to-image translation model to generate RGB images to better understand the quality of the synthesized semantic maps. Finally, we showcase several real-world image-editing scenarios including removal, insertion, and replacement, which enables better user-control over the generated process.
\vskip \secmargin
%------------------------------------------------------------------------
\vskip \secmargin
\section{Acknowledgement}
\vskip \secmargin
This work is supported in part by the NSF CAREER Grant $\#1149783$, MOST 108-2634-F-007-016-, and MOST 109-2634-F-007-016-.

\clearpage
\bibliographystyle{splncs04}
\bibliography{egbib}

% supp
\clearpage
%\clearpage

\begin{appendix}
%\appendix

% \clearpage % For skip to next page
\section{Overview}
We provide more qualitative and quantitative results in \secref{quan_results} and \secref{qual_results}. For a quick walkthrough of this paper, please also check the video provided in the zip file.

\section{Quantitative Results}
\label{sec:quan_results}
%\subsection{Compared with ablated models}
We measure the ``compatibility error'' and ``reconstruction error'' on the ablated models in \tabref{quan_supp_compat_recon}.
Adopting LSTM in the iterative prediction process helps the model to generate more compatible and reasonable shapes, which greatly improves the performance on the compatibility and the reconstruction.
Leveraging the learned shape prior further enhances the results since the distribution varies a lot when the input contains a different combination of classes and each shape of the class has variations.
\tabref{quan_supp_compat_recon} shows the necessity of the proposed model’s architecture
design.

\noindent\textbf{Label-set length analysis.}
We compute FID on HumanParsing with different length of label-sets to analyze its effect on the proposed method in \tabref{quan_label_length}. %Table~\ref{tab:quan_label_length}.
The results show that the larger the length, the larger the FID.

\noindent\textbf{Randomized prediction order.}
To further analyze how the prediction order will affect the performance, we randomly shuffle the prediction order during training and evaluate the FID in \tabref{quan_random_order}.
\section{Qualitative Results}
\label{sec:qual_results}
\begin{figure}[t]
    \centering
    \includegraphics[width=\linewidth]{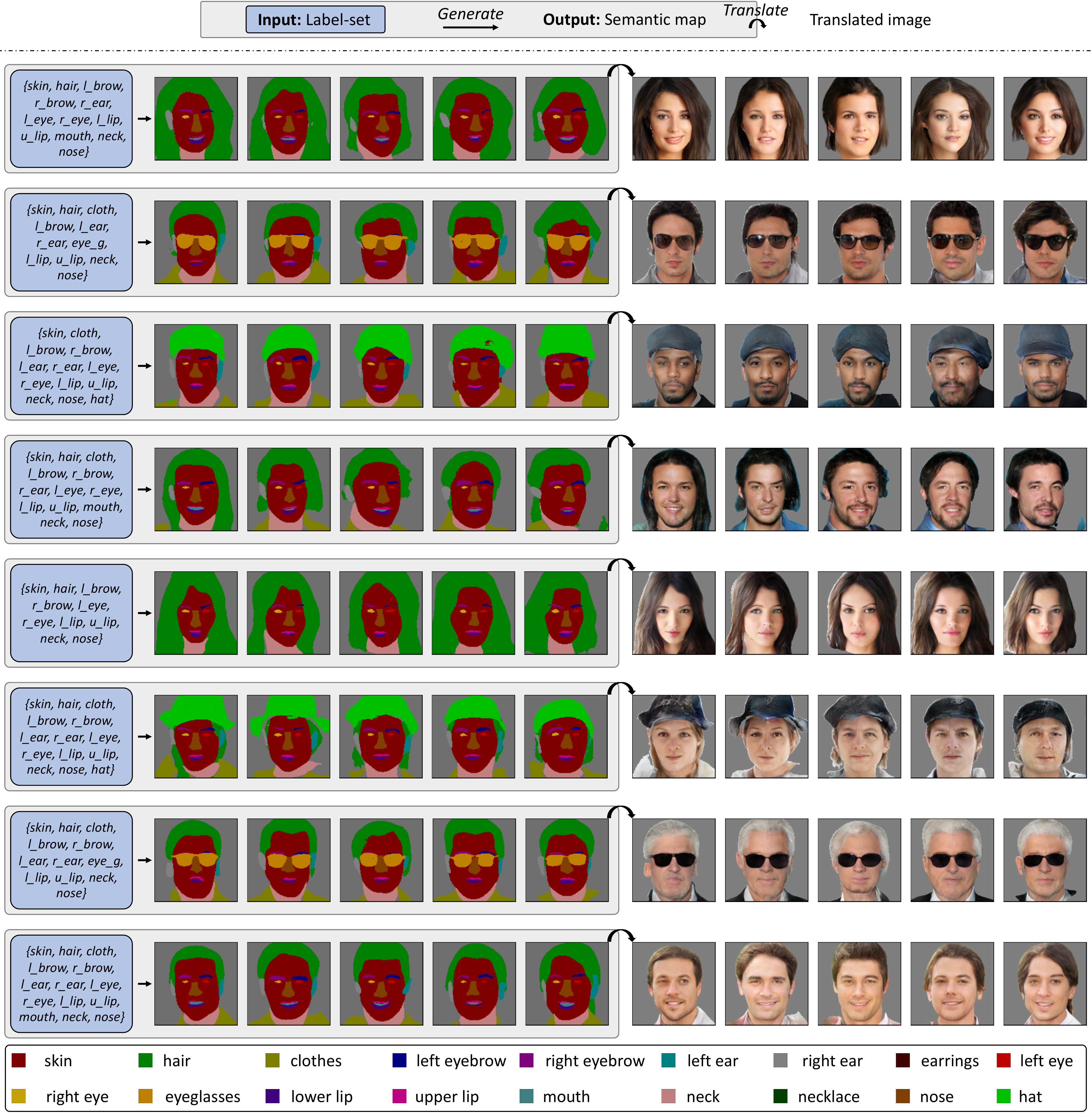}
    
    \caption{\textbf{Multi-modality.}
    We demonstrate the ability of the proposed model to generate diverse results given a label-set on the CelebAMask-HQ dataset.} 
    \label{fig:supp_multi_modal_face} 
    \vspace{\figmargin}
\end{figure}

\begin{figure}[t]
    \centering
    \includegraphics[width=\linewidth]{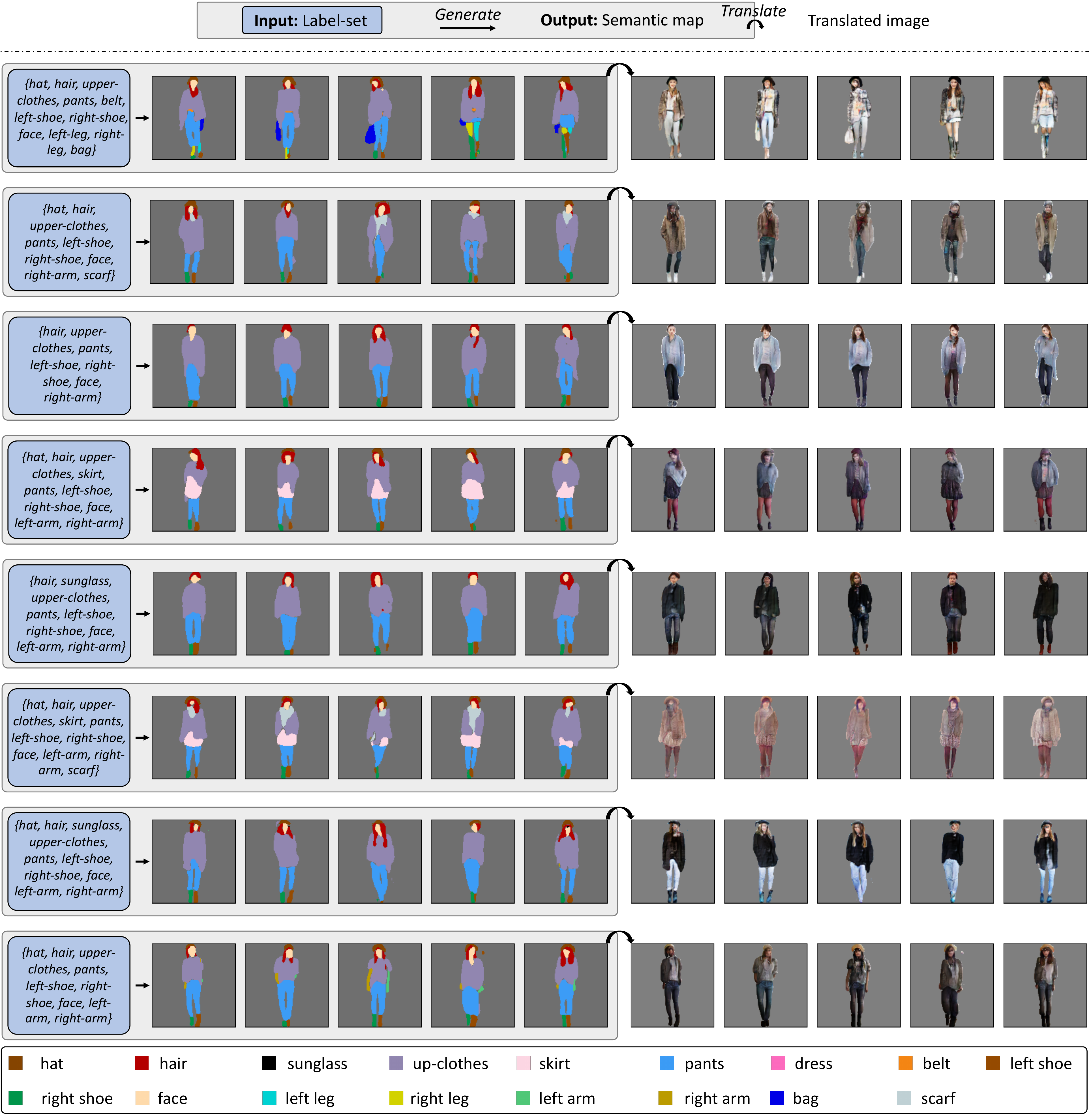}
    
    \caption{\textbf{Multi-modality.}
    We demonstrate the diverse generation results given a label-set on the HumanParsing dataset.}
    \label{fig:supp_multi_modal_hp} 
    \vspace{\figmargin}
\end{figure}

\begin{figure}[t]
    \centering
    \includegraphics[width=\linewidth]{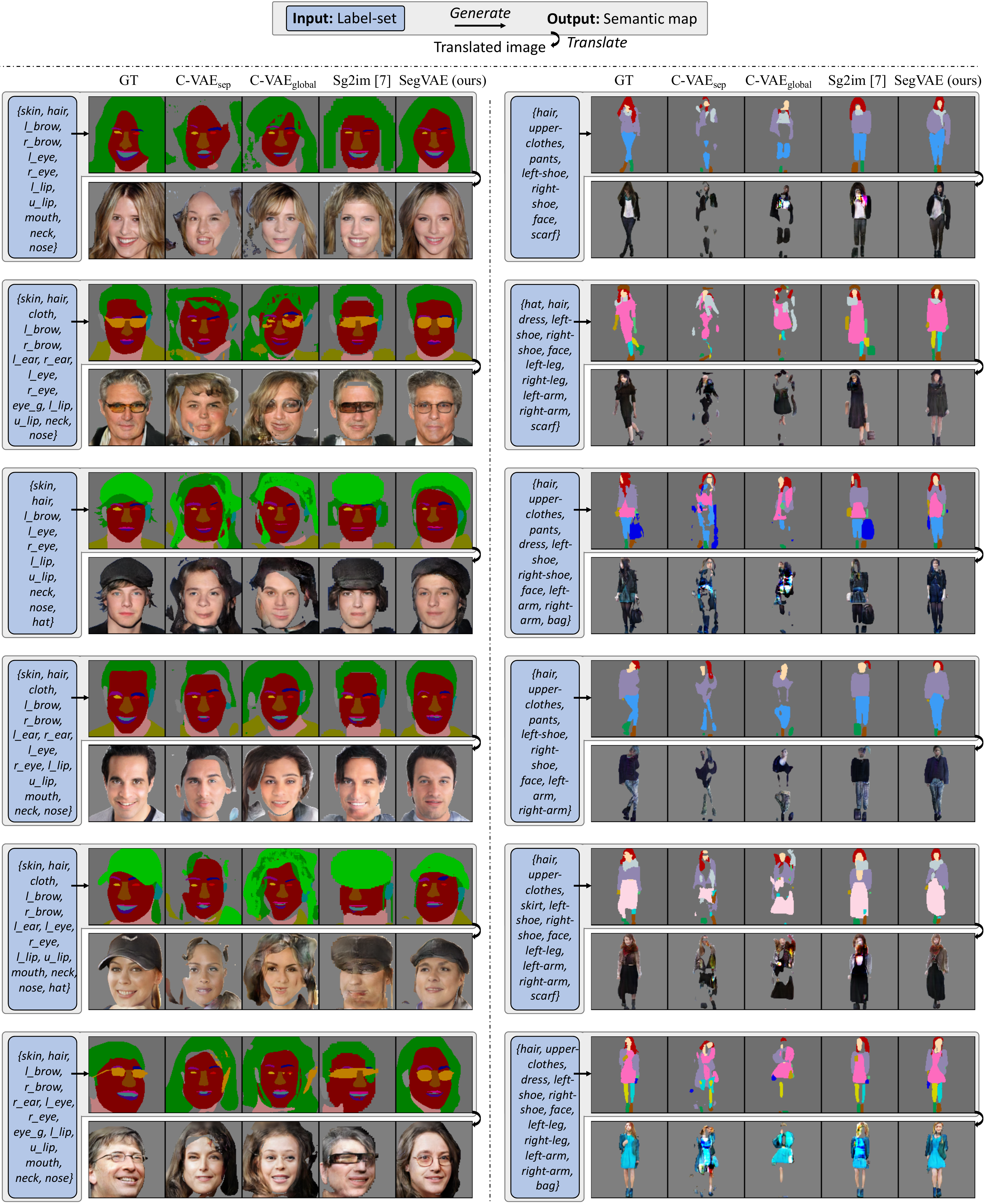}
    \vspace{-1.2em}
    \caption{\textbf{Qualitative comparison.} 
    We present the generated semantic maps given a label-set on the CelebAMask-HQ (left) and the HumanParsing (right) datasets.
    The proposed model generates images with better visual quality compared to other methods.
    We also present the translated realistic images via SPADE~\cite{park2019SPADE}.
    Please refer to \figref{supp_multi_modal_face} and \figref{supp_multi_modal_hp} for the color mapping for each category.}
    \label{fig:supp_comp_baselines} 
    \vspace{\figmargin}
\end{figure}

\begin{figure}[t!]
    \centering
    \includegraphics[width=\linewidth]{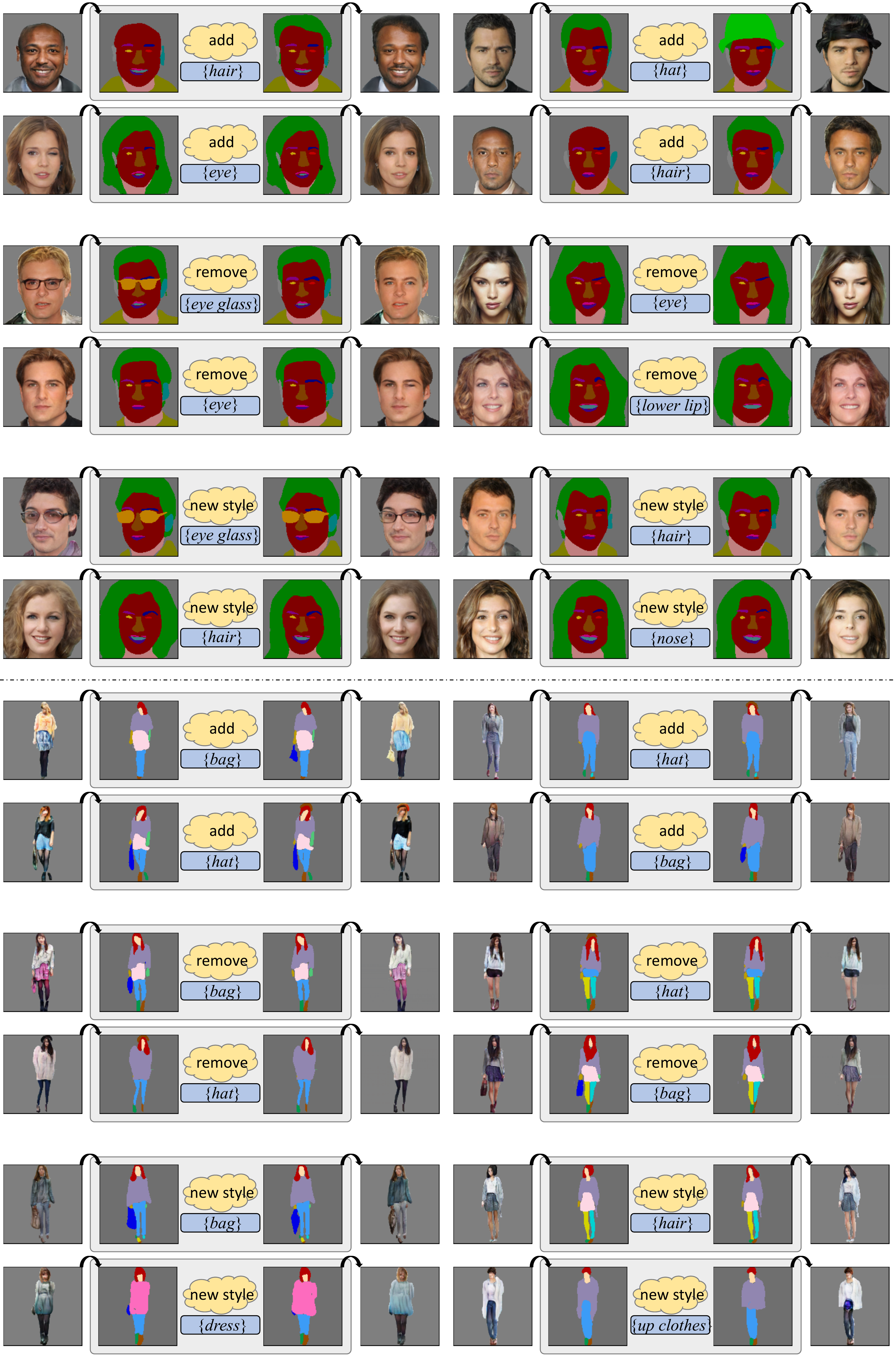}
 
    \caption{\textbf{Editing.}
    We present three real-world image editing applications: \textit{add}, \textit{remove}, and \textit{new style}, on the CelebAMask-HQ and HumanParsing datasets.
    SegVAE enables flexible and intuitive control over the generated outputs.
    } 
    \label{fig:supp_seg_editing} 
    \vspace{\figmargin}
    \vspace{-1.2em}
\end{figure}

We present more generated results in this section. 
For multimodal generation, please see \figref{supp_multi_modal_face} and \figref{supp_multi_modal_hp}. 
For comparison with baselines and the existing method, please refer to \figref{supp_comp_baselines}. 
Finally, more editing results are demonstrated in \figref{supp_seg_editing}.

\begin{figure}[t]
    \centering
    \includegraphics[width=\linewidth]{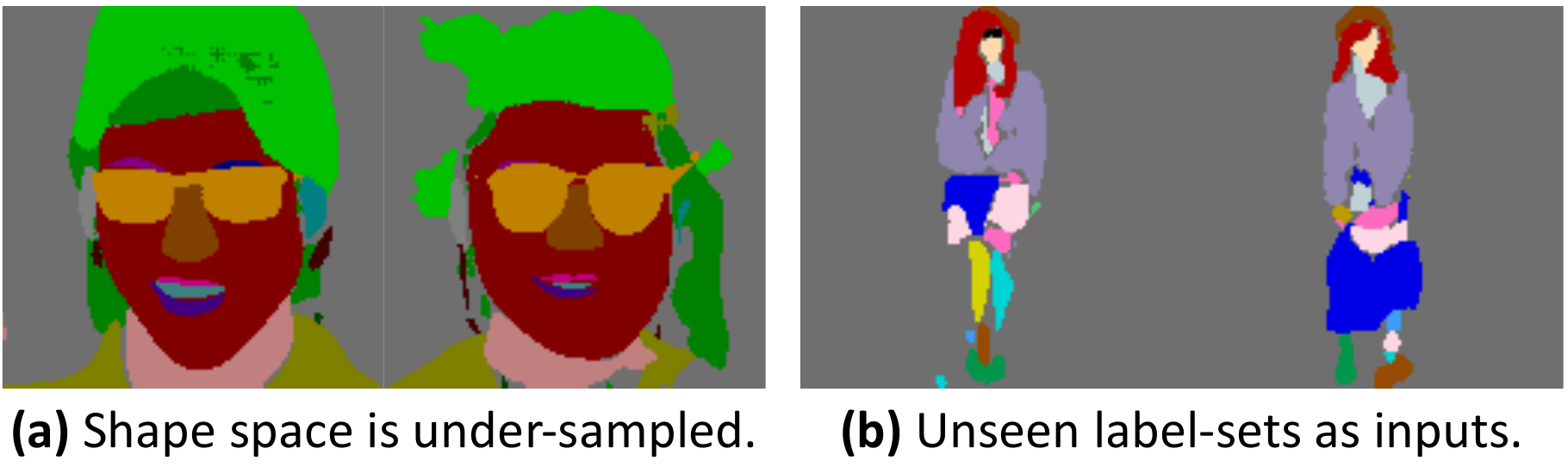}
    
    \caption{\textbf{Failure cases.}
    We demonstrate two typical cases in two datasets: (a) the shape space is under-sampled, (b) the input label-set is unseen during training.}
    \label{fig:supp_failure} 
    \vspace{\figmargin}
    \vspace{-1.2em}
\end{figure}

\noindent\textbf{Failure cases.}
The proposed method has the following limitations.
First, SegVAE fails when the shape vector of a certain class is located in an under-sampled space. For example, \textit{hat} in the CelebAMask-HQ as shown in \figref{supp_failure}~(a).
Second, the label-set combination is unseen.
\figref{supp_failure}~(b) shows the output of the proposed model when all classes are used in a label-set.
SegVAE has difficulty to handle this input since HumanParsing do not contain this example.

\begin{table}[h]
\setlength{\tabcolsep}{2.5pt}
\caption{\tb{Compatibility error and reconstruction error.} 
We train a shape predictor to measure the compatibility error (abbreviated as ``Compat. err.'') over the generated shapes for all methods. We also train an auto-encoder to measure the quality of our generated results by calculating the reconstruction error (denoted as ``Recon. err.'').}
\centering
\begin{tabular}{l  cc cc} 
    \toprule
     & \multicolumn{2}{c}{HumanParsing} & \multicolumn{2}{c}{CelebAMask-HQ} \\
    \cmidrule(r){2-3} \cmidrule(r){4-5}
    {Method} & {Compat. err.~$\shortdownarrow$} & {Recon. err.~$\shortdownarrow$} & {Compat. err.~$\shortdownarrow$} & {Recon. err.~$\shortdownarrow$} \\
    \midrule
    $\text{Ours w/o LSTM}$ & $.7534^{\rpm .0249}$ & $.6808^{\rpm .009}$ & $.0987^{\rpm .0025}$ & $.1144^{\rpm.002}$ \\
    $\text{Ours w/o Learned Prior}$     & $ .6926^{\rpm .0115}$ & $.5856^{\rpm .010}$ & $.0839^{\rpm .0029}$ & $.1045^{\rpm .003}$ \\
    Ours & $\mathbf{.6174^{\rpm .0147}}$ & $\mathbf{.5663^{\rpm .011}}$ & $\mathbf{.0754^{\rpm .0013}}$ & $\mathbf{.0840^{\rpm.001}}$ \\
    \bottomrule
\end{tabular}
%\vspace{\tabmargin}
\vskip \tabmargin
\label{tab:quan_supp_compat_recon}
\end{table}
%\begin{table}[t]
\begin{table}[h]
\caption{\textbf{Label-set length analysis}. We compute FID on HumanParsing with different length of label-sets.
The results show that the larger the length, the larger the FID.
}
\centering
\small
\begin{tabular}{l  cc} 
    \toprule
     %& \multicolumn{2}{c}{HumanParsing} \\
    %\cmidrule(r){2-3}
    Length & {FID}$\shortdownarrow$ \\
    \midrule
    $10$ & 39.2692 \\
    $11$ & 39.5351 \\
    $12$ & 39.9063 \\
    $13$ & 41.0591 \\
    $14$ & 44.2884 \\
    \bottomrule
\end{tabular}
\label{tab:quan_label_length}
\end{table}
\begin{table}[t]
%\begin{table}[h]
\caption{\textbf{Prediction order analysis}. 
}
\centering
\small
\begin{tabular}{l  cc} 
    \toprule
     %& \multicolumn{2}{c}{HumanParsing} \\
    %\cmidrule(r){2-3}
    {Order}  (HumanParsing) & {FID}$\shortdownarrow$ & {Diversity}$\shortuparrow$ \\
    \midrule
    $\text{1}~\textit{Body} \rightarrow \textit{Clothes} \rightarrow \textit{Accessories}~\text{(Ours)}$ & $\mathbf{39.6496^{\rpm .3543}}$ & $\mathbf{.2072^{\rpm .053}}$ \\
    $\text{2}~\textit{Clothes} \rightarrow \textit{Body} \rightarrow \textit{Accessories}$ & $39.9008^{\rpm .5263}$ & $.2062^{\rpm .0494}$  \\
    $\text{3}~\textit{Accessories} \rightarrow \textit{Body} \rightarrow \textit{Clothes}$ & $40.2909^{\rpm .2195}$ & $.2043^{\rpm .0521}$ \\
    \textit{Random order} & $ 52.6728^{\rpm .3865}$& $.1714^{\rpm 0.044}$\\
    \bottomrule
\end{tabular}
% \\
% \vspace{1em}
\begin{tabular}{l  cc} 
    \toprule
    % & \multicolumn{2}{c}{CelebAMask-HQ} \\
    %\cmidrule(r){2-3}
    {Order} (CelebAMask-HQ) & {FID}$\shortdownarrow$ & {Diversity}$\shortuparrow$ \\
    \midrule
    $\text{1}~\textit{Face} \rightarrow \textit{Face features} \rightarrow \textit{Accessories}~\text{(Ours)}$ & $\mathbf{28.8221^{\rpm .2732}}$ & $\mathbf{.1575^{\rpm .043}}$ \\
    $\text{2}~\textit{Face features} \rightarrow \textit{Face} \rightarrow \textit{Accessories}$ & $30.6547^{\rpm .1267}$ & $.1517^{\rpm .0376}$ \\
    $\text{3}~\textit{Accessories} \rightarrow \textit{Face} \rightarrow \textit{Face features}$ & $32.0325^{\rpm .1294}$ & $.1489^{\rpm .0363}$ \\
    \textit{Random order} & $ 31.1238^{\rpm .2088}$ & $.1529^{\rpm 0.042}$\\
    \bottomrule
\end{tabular}
%\vspace{\tabmargin}
\vskip \tabmargin
\label{tab:quan_random_order}
\end{table}

% figures

\end{appendix}

\end{document}